\documentclass[letterpaper]{article} %
\usepackage{aaai2026}  %
\usepackage{times}  %
\usepackage{helvet}  %
\usepackage{courier}  %
\usepackage[hyphens]{url}  %
\usepackage{graphicx} %
\usepackage{natbib}  %
\usepackage{caption} %
\usepackage{amssymb}
\usepackage{amsmath}
\usepackage{graphicx}
\usepackage{subcaption}
\usepackage{booktabs}
\usepackage{multirow}
\usepackage{pifont}

\usepackage{float}
\usepackage{newfloat}
\usepackage[linesnumbered,ruled,vlined]{algorithm2e}
\usepackage{newfloat}
\usepackage{listings}
\DeclareCaptionStyle{ruled}{labelfont=normalfont,labelsep=colon,strut=off} %
\floatstyle{ruled}
\newfloat{listing}{tb}{lst}{}
\floatname{listing}{Listing}
\title{Privacy Auditing of Multi-domain Graph Pre-trained Model \\ under Membership Inference Attacks}

\author {
    Jiai Luo\textsuperscript{\rm 1},
    Qingyun Sun\textsuperscript{\rm 1},
    Yuecen Wei\textsuperscript{\rm 2},
    Haonan Yuan\textsuperscript{\rm 1},\\
    Xingcheng Fu\textsuperscript{\rm 3},
    Jianxin Li\textsuperscript{\rm 1}\thanks{Corresponding Author}
}
\affiliations {
    \textsuperscript{\rm 1}SKLCCSE, School of Computer Science and Engineering, Beihang University, Beijing, China\\
    \textsuperscript{\rm 2}School of Software, Beihang University, China\\
    \textsuperscript{\rm 3}Key Lab of Education Blockchain and Intelligent Technology, Ministry of Education, Guangxi Normal University, China\\
    \{luojy, sunqy, yuanhn, lijx\}@buaa.edu.cn, 
    weiyc@buaa.edu.cn, 
    fuxc@gxnu.edu.cn
}

\usepackage{bibentry}

\begin{document}

\maketitle

\newcommand{\Model}{\textsc{Mgp-Mia}}
\begin{abstract}

Multi-domain graph pre-training has emerged as a pivotal technique in developing graph foundation models. 
While it greatly improves the generalization of graph neural networks, its privacy risks under membership inference attacks (MIAs), which aim to identify whether a specific instance was used in training (member), remain largely unexplored.
However, effectively conducting MIAs against multi-domain graph pre-trained models is a significant challenge due to:
\textit{(\romannumeral1) Enhanced Generalization Capability}: Multi-domain pre-training reduces the overfitting characteristics commonly exploited by MIAs.
\textit{(\romannumeral2) Unrepresentative Shadow Datasets}: Diverse training graphs hinder the obtaining of reliable shadow graphs.
\textit{(\romannumeral3) Weakened Membership Signals}: Embedding-based outputs offer less informative cues than logits for MIAs.
To tackle these challenges, we propose \textbf{\underline{\Model}}\footnote{Code: \url{https://github.com/RingBDStack/MGP-MIA}.}, a novel framework for \textbf{\underline{M}}embership \textbf{\underline{I}}nference \textbf{\underline{A}}ttacks against \textbf{\underline{M}}ulti-domain \textbf{\underline{G}}raph \textbf{\underline{P}}re-trained models.
Specifically, we first propose a membership signal amplification mechanism that amplifies the overfitting characteristics of target models via machine unlearning. 
We then design an incremental shadow model construction mechanism that builds a reliable shadow model with limited shadow graphs via incremental learning.
Finally, we introduce a similarity-based inference mechanism that identifies members based on their similarity to positive and negative samples.
Extensive experiments demonstrate the effectiveness of our proposed \Model\ and reveal the privacy risks of multi-domain graph pre-training.
\end{abstract}

\section{Introduction}
Multi-domain graph pre-training~\cite{zhao2024all,yu2024text,yu2025samgpt,wang2025multi,yuan2025how} has emerged as a critical technique for developing graph foundation models~\cite{zhao2025survey,shi2024graph,mao2024graph,liu2025graph,wang2025graph}. By pre-training Graph Neural Networks (GNNs) across graphs from diverse domains using self-supervised learning paradigms such as link prediction~\cite{zhang2018link} and contrastive learning~\cite{you2020graph}, multi-domain graph pre-training enables the pre-trained GNNs to capture transferable structural and semantic patterns, thereby improving the generalization across a wide range of downstream tasks.

\begin{figure}[t]
  \centering
  \includegraphics[width=\linewidth]{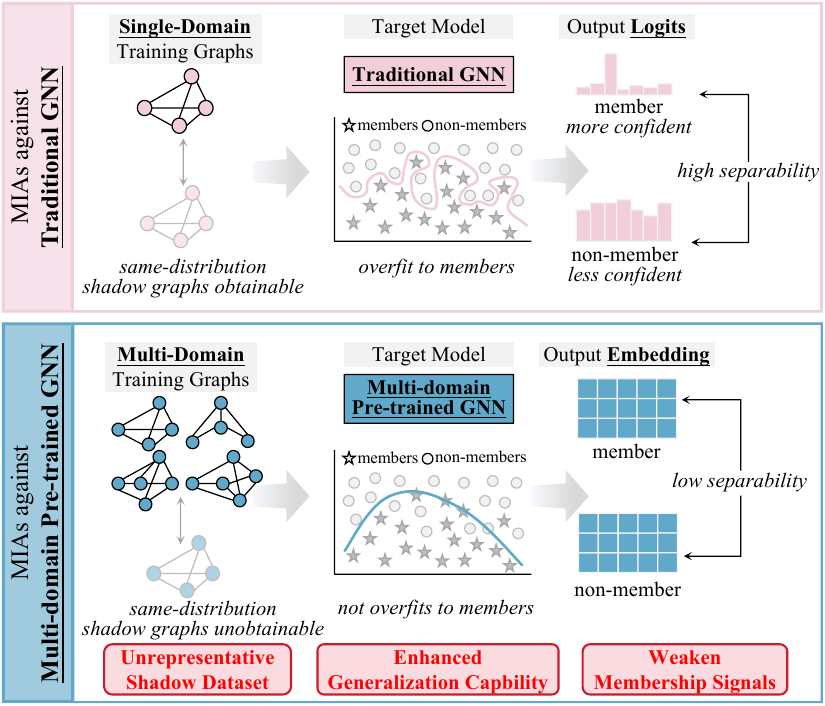}
  \caption{A comparison between graph MIAs against Traditional GNNs and Multi-domain Pre-trained GNNs.}
  \label{fig:motivation}
\end{figure}

Although multi-domain graph pre-training significantly improves graph learning performance~\cite{zhao2024all,yu2024text,yu2025samgpt,wang2025multi,yuan2025how}, its privacy vulnerabilities under Membership Inference Attacks (MIAs) remain largely unexplored.
MIAs are privacy attacks that aim to determine whether a specific data sample was included in a model’s training set~\cite{olatunji2021membership,he2021node,wei2024poincare,wei2025prompt,wang2025out,song2025mitigating,luo2025robust}. 
When developers publish pre-trained models on open-source platforms to promote reuse and enable downstream users to build their own graph foundation models, they may inadvertently expose these models to privacy risks. 
Adversaries can exploit the released models to infer the presence of sensitive records in the training data, leading to the disclosure of user-related information.

Typical MIAs involve two stages: the attacker first trains shadow models on shadow data from the target distribution to mimic the target model, then queries the shadow models with member and non-member samples in the shadow data to collect prediction logits, from which it learns a rule to infer whether a target sample was in training, based on the fact that models often give higher confidence to members than non-members.
However, as shown in Figure~\ref{fig:motivation}, performing effective MIAs against multi-domain graph pre-trained models remains significantly challenging due to three key factors:
\textbf{(\romannumeral1) Improved Generalization Capability:} 
Multi-domain graph pre-training can capture transferable knowledge that improves generalization, thereby mitigating the overfitting signals commonly exploited by MIAs.
\textbf{(\romannumeral2) Unrepresentative Shadow Datasets:}
Multi-domain pre-training scenarios involve diverse training graphs from multiple domains, making it difficult to obtain domain-aligned shadow graphs and thus hindering the construction of a reliable and effective shadow model for inference attack.
\textbf{(\romannumeral3) Weakened Membership Signals:} 
Since pre-trained encoders output embeddings rather than logits, the resulting representations carry weaker overfitting signals, making it harder to distinguish members from non-members.

The aforementioned challenges limit the effectiveness of existing graph MIAs, as further discussed in Section~\ref{sec:pre_analysis}. 
To address these limitations, we propose \textbf{\underline{\Model}}, a novel framework for \textbf{\underline{M}}embership \textbf{\underline{I}}nference \textbf{\underline{A}}ttacks targeting \textbf{\underline{M}}ulti-domain \textbf{\underline{G}}raph \textbf{\underline{P}}re-trained models.
Specifically, \textit{to amplify membership signals in the target model and counteract the generalization effect introduced by multi-domain pre-training}, we propose a membership signal amplification mechanism that increases the overfitting degree of the target model via machine unlearning, motivated by the observation that machine unlearning can release parameter capacity and induce stronger memorization on the remaining data.
\textit{To construct a reliable shadow model without access to domain-aligned shadow data}, we design an incremental shadow model construction mechanism. Instead of training a shadow model from scratch, this mechanism constructs the shadow model by fine-tuning the target model using parameter regularization in an incremental learning manner, allowing it to better approximate the membership inference characteristics of the target model only with a limited shadow graph. 
\textit{To extract membership signals from output embeddings in the absence of explicit cues like logits}, we introduce a similarity-based inference mechanism. This mechanism constructs attack features by measuring the similarity between a target node and its positive and negative samples, and infers membership status based on these similarity patterns. This design is motivated by the fact that self-supervised pre-training paradigms generally encourage embeddings to move closer to positive samples and farther from negative ones.
We conduct extensive experiments targeting the representative multi-domain graph pre-training methods and demonstrate the superior performance of \Model.

In summary, our main contributions are as follows:

\begin{itemize}
    \item We audit the privacy leakage of multi-domain graph pre-trained models under Membership Inference Attacks (MIAs) and propose a novel approach named \Model, for performing such attacks. To the best of our knowledge, this is the first work to investigate this problem.
    \item We propose a membership signal amplification mechanism to counteract the generalization effects of multi-domain pre-training.
    We design an incremental shadow model construction mechanism to build reliable shadow models without domain-aligned shadow data.
    We introduce a similarity-based inference mechanism to extract membership signals from output embeddings.
    \item Extensive experiments against multi-domain graph pre-training models demonstrate the effectiveness of \Model\ and reveal their privacy vulnerabilities to MIAs.
\end{itemize}

\section{Related Work}

\noindent \textbf{Multi-domain Graph Pre-training.} Multi-domain graph pre-training serves as the basis for graph foundation models~\cite{shi2024graph,mao2024graph,liu2025graph,wang2025graph} by applying self-supervised learning to graphs from diverse domains~\cite{zhao2024all,yu2024text,yu2025samgpt,yuan2025how,wang2025multi}, facilitating the extraction of transferable knowledge.
Existing approaches adopt either graph contrastive learning~\cite{you2020graph} or link prediction~\cite{zhang2018link} objectives.
In contrastive learning, GCOPE~\cite{zhao2024all} adds virtual nodes to link domains and optimizes them dynamically during training, while SAMGPT~\cite{yu2025samgpt} enhances structural consistency using structure tokens to unify message aggregation. MDGFM~\cite{wang2025multi} adaptively balances features and topology, refining graphs to reduce noise and align structures.
In link prediction, MDGPT~\cite{yu2024text} uses domain tokens to align semantic features, and BRIDGE~\cite{yuan2025how} introduces a domain aligner to extract shared representations and suppress noise.

\noindent\textbf{Graph Membership Inference Attack.} 
Membership inference attacks (MIAs) aim to determine whether a specific data instance was used during the training of a target model~\cite{hu2022membership}. 
\citet{he2021node} and \citet{olatunji2021membership} first extend MIAs to GNNs by incorporating structural information into both target and shadow models. 
ProIA~\cite{wei2025prompt} enhanced inference performance by introducing prompt-based techniques to augment the attack model’s background knowledge. 
Meanwhile, \textsc{GCL-Leak}~\cite{wang2024gcl} is the first graph MIA against contrastive learning, but its focus on the federated setting with access to the training process makes it inapplicable to our scenario.
Subsequent work \citet{zhang2022inference} and \citet{wu2021adapting} targeted whole-graph membership, and \citet{wang2022group} conducted a systematic study on inferring the membership of particular groups of nodes and links.
To operate under more constrained adversarial settings, \citet{conti2022label} and \citet{dai2025graph} propose label-only MIAs.

\begin{figure*}[!t]
  \centering
  \begin{minipage}[t]{0.48\linewidth}
    \centering
    \includegraphics[width=\linewidth]{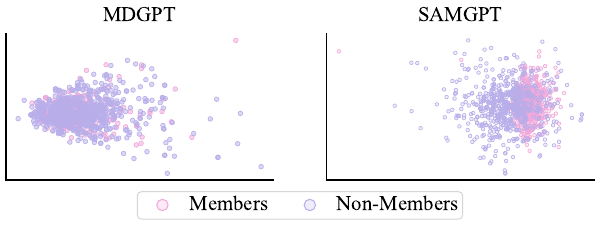}
    \caption{Separability analysis of output node embeddings.}\label{fig:ana_sparability}
  \end{minipage}
  \begin{minipage}[t]{0.48\linewidth}
    \centering
    \includegraphics[width=\linewidth]{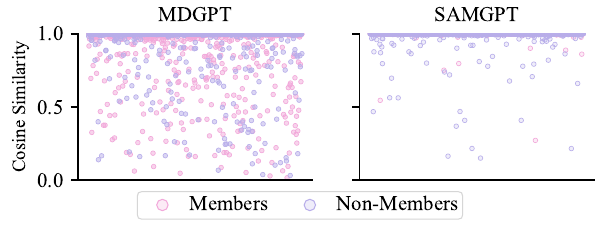}
    \caption{Robustness analysis of output node embeddings.}\label{fig:ana_robustness}
  \end{minipage}
\end{figure*}

\section{Problem Formulation}
The goal of MIAs is to infer whether a given instance was part of the training set $\mathcal{D}_{\text{Train}}$ of a target model $\mathcal{F}_{\text{Target}}$. Typically, the adversary uses a shadow dataset $\mathcal{D}_{\text{Shadow}}$ from the same distribution as $\mathcal{D}_{\text{Train}}$ to train a shadow model $\mathcal{F}_{\text{Shadow}}$ that mimics the target. Based on the outputs of $\mathcal{F}_{\text{Shadow}}$ on $\mathcal{D}_{\text{Train}}$ and $\mathcal{D}_{\text{Test}}$, the adversary constructs an attack model $\mathcal{F}_{\text{Attack}}$ to infer whether a given instance is a member.

\subsubsection{Attacker's Goal.}
We focus on the node-level membership inference attacks, following prior work~\cite{hu2022membership, olatunji2021membership, wei2025prompt}. Given a target GNN $\mathcal{F}_{\text{Train}}$ pre-trained across graphs $\{\mathcal{G}_{i}\}_{i=1}^{m}$ from multiple domains, the goal is to determine whether a specific node $v$ was included in the training set. Formally, the attacker aims to learn an attack model $\mathcal{F}_{\text{Attack}}$ that maximizes the expected accuracy of membership prediction as follows:
\begin{equation}
\max_{\mathcal{F}_{\text{Attack}}} \mathbb{E}_{v} \left[ \mathbb{I} \left( \mathcal{F}_{\text{Attack}}(v, \mathcal{F}_{\text{Target}}) = y_v \right) \right],
\end{equation}
where $y_v \in \{0, 1\}$ indicates whether node $v$ belongs to the training set of $\mathcal{F}_{\text{Target}}$ (1 for member, 0 for non-member), and $\mathbb{I}(\cdot)$ denotes the indicator function.

\subsubsection{Attacker's Knowledge.}
We assume a white box adversary with full access to the target model $\mathcal{F}_{\text{Target}}$, including its architecture, parameters, and training algorithm. 
However, the attacker does not have access to the original training process or the complete multi-domain training dataset. Instead, the attacker is assumed to have a shadow graph from one domain that shares a similar distribution with the target node $v$ under inference.
\textit{This setting reflects practical scenarios where costly pre-trained models are publicly released to support downstream GFM development~\cite{liu2025graph}, potentially exposing privacy vulnerabilities.}

\section{Pre-attack Analysis}\label{sec:pre_analysis}
In this section, we analyze \textit{why existing graph MIAs are not applicable to multi-domain graph pre-trained models}. Existing methods can be categorized into two types: confidence-based and stability-based. Confidence-based attacks exploit confidence scores in the prediction logits, while Stability-based attacks assess the stability of predicted labels. Both rely on the observation that member instances typically yield more confident (for confidence-based) or more stable (for stability-based) predictions than non-members.
However, the outputs of multi-domain graph pre-trained models are node embeddings. To assess whether these embeddings exhibit separability similar to prediction logits or stability similar to output labels for members, we conduct toy experiments on the Cora~\cite{yang2016revisiting} dataset using MDGPT~\cite{yu2024text} (link prediction) and SAMGPT~\cite{yu2025samgpt} (contrastive learning) as representative victims. We summarize two key observations below (More experimental details are provided in Appendix~\ref{appendix:pre_attack}):

\textit{\ding{72}Takeaway 1: Embeddings are weakly separable between members and non-members.}
To assess whether the embeddings exhibit separability between members and non-members similar to the logits, we visualize the embeddings using PCA~\cite{abdi2010principal} in Figure~\ref{fig:ana_sparability}. As shown, the embeddings themselves are not clearly separable.

\textit{\ding{72}Takeaway 2: Embeddings are not more stable for members.}
We perturb the graph edges and visualize the similarity between original and perturbed embeddings. As shown in Figure~\ref{fig:ana_robustness}, member embeddings are not more stable under perturbation, particularly in link-prediction-based methods.

In summary, both the weak separability and the lack of enhanced stability for member nodes in the output embeddings clearly highlight the inherent limitations of directly applying existing graph membership inference attacks against multi-domain graph pre-trained models.

\begin{figure*}[t]
  \centering
  \includegraphics[width=\textwidth]{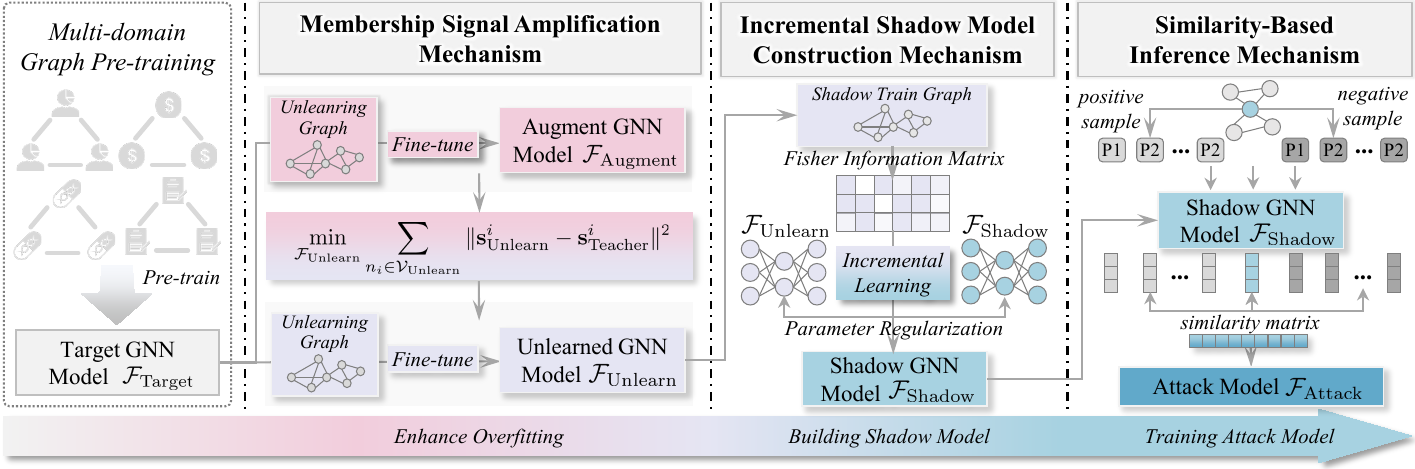}
  \caption{Overview of \Model. 
  The membership signal amplification mechanism first leverages machine unlearning to enhance overfitting.
  The shadow model is then built from the unlearned model using incremental learning. Finally, the attack features are derived from similarities between each sample and its positive and negative counterparts to train the attack model.}
  \label{fig:framework}
\end{figure*}

\section{\Model}
In this section, we present \textbf{\underline{\Model}}, a novel framework for \textbf{\underline{M}}embership \textbf{\underline{I}}nference \textbf{\underline{A}}ttacks against \textbf{\underline{M}}ulti-domain \textbf{\underline{G}}raph \textbf{\underline{P}}re-trained models.
As illustrated in Figure~\ref{fig:framework}, our proposed \Model\ consists of three key components.
First, the membership signal amplification mechanism enhances membership signals by mitigating the generalization effect of multi-domain training through machine unlearning.
Second, the incremental shadow model construction mechanism employs incremental learning to build a shadow model that replicates the overfitting behavior of the target model only with the limited shadow dataset.
Finally, the similarity-based inference mechanism constructs attack features by measuring the similarity between a target sample and its positive and negative samples, which are then used to train an attack model for membership inference.

\subsection{Membership Signal Amplification Mechanism}
\label{sec:unlearning}
Multi-domain pre-training trains GNNs on multiple graphs to learn transferable knowledge that generalizes across diverse downstream tasks and domains. However, the resulting improved generalization reduces the model's overfitting behavior, which in turn weakens the critical membership signals and makes effective attacks much more difficult.

To address this challenge, we propose a membership signal amplification mechanism that enhances membership signals through machine unlearning. 
Machine unlearning is a privacy-preserving strategy that removes the influence of specific data instances from a trained model, making it behave as if those instances were never used during training~\cite{chen2022graph}.
However, prior work has shown that inexact machine unlearning can lead to stronger overfitting on the remaining samples as it erases specific learned information and reallocates model capacity~\cite{hayes2025inexact}. 
Motivated by this phenomenon, we repurpose this privacy protection strategy as a tool to help us enhance the effectiveness of membership inference privacy attacks.
While vanilla machine unlearning methods that do not require access to the remaining data typically perform unlearning through gradient ascent~\cite{jang2022knowledge,rashid2025forget}. 
Directly applying this approach ignores the diverse characteristics of different nodes and can severely degrade model utility.
Inspired by recent advances in unlearning for large language models~\cite{wang2025balancing}, we propose a selective and controllable unlearning strategy that employs an argument model to identify disproportionately memorized nodes and guide the model to suppress overfitted components while maintaining the utility of the unlearned model.

Specifically, we randomly extract a subgraph $\mathcal{G}_{\text{Unlearn}}$ from the given shadow graph $\mathcal{G}_{\text{Shadow}}$ as the target for inexact machine unlearning. The target model $\mathcal{F}_{\text{Target}}$ is fine-tuned on $\mathcal{G}_{\text{Unlearn}}$ for a few epochs to produce the augment model $\mathcal{F}_{\text{Augment}}$. 
We then compare the output embeddings of $\mathcal{F}_{\text{Target}}$ and $\mathcal{F}_{\text{Augment}}$ to evaluate how the similarity between each node and its positive and negative neighbors changes under the two models.
Specifically, for a given node $v_i$, we define the similarity score vector as follows:
\begin{equation}\label{eq:similarity}
\mathbf{s}^{i} = \left[
\begin{aligned}
&\text{sim}(\mathbf{h}_i, \mathbf{h}_{i_1^+}), \ldots, \text{sim}(\mathbf{h}_i, \mathbf{h}_{i_P^+}), \\
&\text{sim}(\mathbf{h}_i, \mathbf{h}_{i_1^-}), \ldots, \text{sim}(\mathbf{h}_i, \mathbf{h}_{i_N^-})
\end{aligned}
\right],
\end{equation}
where $\mathbf{h}_{i}$ is the out embedding of node $i$ produced by model, $\{i_{p}^{+}\}_{p=1}^{P}$ are its positive samples, $\{i_{n}^{-}\}_{n=1}^{N}$ are its negative samples, and $\operatorname{sim}(\cdot,\cdot)$ denotes the cosine similarity function.
Let $\mathbf{s}_{\text{Target}}^{i}$ and $\mathbf{s}_{\text{Augment}}^{i}$ denote the similarity between the output embeddings of node $i$ and its positive and negative samples under the target and augment models, respectively. 
The difference $(\mathbf{s}_{\text{Target}}^{i} - \mathbf{s}_{\text{Augment}}^{i})$ captures the memorization sensitivity of node $i$, indicating the extent to which its associated knowledge should be unlearned.
This difference is used to compute the teacher similarity score $\mathbf{s}_{\text{Teacher}}^{i}$, which serves as the guiding signal for the unlearning process:
\begin{equation}
\mathbf{s}_{\text{Teacher}}^{i}=\mathbf{s}_{\text{Target}}^{i} - \lambda \cdot (\mathbf{s}_{\text{Target}}^{i} - \mathbf{s}_{\text{Augment}}^{i}),
\end{equation}
where $\lambda$ is the hyperparameter controlling the strength of machine unlearning. 
Finally, we fine-tune the target model $\mathcal{F}_{\text{target}}$ to minimize the deviation between the student similarity score $\mathbf{s}_{\text{Unlearn}}^{i}$ and the teacher score $\mathbf{s}_{\text{Teacher}}^{i}$, resulting in the unlearned model $\mathcal{F}_{\text{Unlearn}}$:
\begin{equation}
    \min_{\mathcal{F}_{\text{Unlearn}}} \sum_{n_{i} \in \mathcal{V}_{\text{Unlearn}}} \Vert \mathbf{s}_{\text{Unlearn}}^{i} - \mathbf{s}_{\text{Teacher}}^{i} \Vert^2,
\end{equation}
where $\mathcal{V}_{\text{unlearn}}$ is the node set of the unlearning graph $\mathcal{G}_{\text{Unlearn}}$. 
The unlearned model $\mathcal{F}_{\text{Unlearn}}$ releases the model capacity previously occupied, allowing the remaining data to be memorized more strongly and thereby enhancing overfitting.

\subsection{Incremental Shadow Model Construction Mechanism}
\label{sec:incremental}
The shadow model is a surrogate trained by an attacker to mimic the target model’s behavior using data from a distribution similar to the target model’s training set. 
As the attacker controls its training process, each instance can be labeled as a member or non-member. 
By recording the shadow model’s outputs with these labels, the attacker builds an attack dataset capturing behavioral differences between seen and unseen data. 
This dataset is then used to train an attack model to infer the real membership of the target model. 

However, in multi-domain graph pre-training, it is impractical to obtain a shadow dataset covering all training domains. A realistic assumption is that the attacker only has access to a shadow graph $\mathcal{G}_{\text{Shadow}}$ from the same domain as the target node $v$ under inference.
To construct a reliable shadow model that mimics the target model’s membership-related behavior using limited shadow data, we propose an incremental shadow model construction mechanism.
Specifically, we randomly split the shadow graph into a training graph $\mathcal{G}_{\text{Shadow}}^{\text{Train}}$ and a test graph $\mathcal{G}_{\text{Shadow}}^{\text{Test}}$.
We use the shadow dataset as a proxy to estimate the Fisher Information Matrix $\mathbf{I}_{\text{Unlearn}}$~\cite{jastrzebski2021catastrophic}, which quantifies the importance of each parameter in $\mathcal{F}_{\text{Unlearn}}$ to the target domain:
\begin{equation}\label{eq:fisher}
    \mathbf{I}_{\text{Unlearn}}(\theta) = \mathbb{E}_{v\sim\mathcal{G}_{\text{Shadow}}^{\text{Train}}}\left[\frac{\partial^{2} \mathcal{L}_{\text{task}}(\mathcal{F}_{\text{Unlearn}};v)}{\partial\theta^{2}}\vert\theta\right],
\end{equation}
where $\theta\in\boldsymbol{\Theta}_{\text{Unlearn}}$ denotes the parameter of the unlearned model $\mathcal{F}_{\text{Unlearn}}$, and $\mathcal{L}_{\text{task}}$ is the loss function associated with the pre-training task. 
We then fine-tune the unlearned model $\mathcal{F}_{\text{Unlearn}}$ on the shadow training graph to obtain the shadow model with the following objective:
\begin{align}
\min_{\boldsymbol{\Theta}_{\text{Shadow}}} \quad
& \sum_{v \in \mathcal{G}_{\text{Shadow}}^{\text{Train}}} 
\mathcal{L}_{\text{task}}(\mathcal{F}_{\text{Shadow}}; v) \nonumber \\
& + \alpha \sum_{i} \mathbf{I}_{\text{Unlearn}}^{(i)} 
\left( \boldsymbol{\Theta}_{\text{Shadow}}^{(i)} - \boldsymbol{\Theta}_{\text{Unlearn}}^{(i)} \right)^2,
\end{align}
where $\mathcal{L}_{\text{task}}$ is the pre-training loss, $\mathbf{I}_{\text{Unlearn}}^{(i)}$ denotes the $i$-th element of the Fisher Information Matrix $\mathbf{I}_{\text{Unlearn}}$, and $\alpha$ controls the regularization strength.

\subsection{Similarity-Based Inference Mechanism}
\label{sec:similarity}
As discussed in Section~\ref{sec:pre_analysis}, the output embeddings from multi-domain graph pre-trained models themselves lack enough membership signals for MIAs. This is because the embeddings are primarily correlated with the intrinsic features of the node and its local neighborhood, rather than with whether the node was included in the model’s training data. 

To extract membership signals from output embeddings, we draw on the principles of self-supervised pre-training, which guide models to encode semantic relationships by pulling positive samples closer and pushing negative ones apart. Following this intuition, we construct attacker features by measuring the embedding similarity between the target node and its associated positive and negative samples.
Specifically, for each target node $v$, we randomly select $m$ positive samples $\{v_{i}^{+}\}_{i=1}^{m}$ and $m$ negative samples $\{v_{i}^{-}\}_{i=1}^{m}$.
In contrastive learning, each $v_{i}^{+}$ is an augmented view of $v$, while $v_{i}^{-}$ is a randomly sampled node unrelated to $v$. 
In link prediction, $v_{i}^{+}$ shares a ground-truth edge with $v$, and $v_{i}^{-}$ does not.
These similarity scores form the feature vector $\mathbf{s}_{v}$ constructed as Eq.~\eqref{eq:similarity} to serve as the attack feature.
To construct the attack dataset $\mathcal{D}_{\text{Attack}}$, we generate similarity-based feature vectors $\mathbf{s}_{v}$ for each node $v$ with shadow model $\mathcal{F}_{\text{Shadow}}$ in the shadow training graph $\mathcal{G}_{\text{Shadow}}^{\text{Train}}$ and shadow testing graphs $\mathcal{G}_{\text{Shadow}}^{\text{Test}}$, respectively. Each node $v$ is associated with a membership label $y_{v}\in\{0,1\}$, where $y_{v}=1$ if $v\in \mathcal{G}_{\text{Shadow}}^{\text{Train}}$ (i.e., a member), and $y_{v}=0$ if $v\in \mathcal{G}_{\text{Shadow}}^{\text{Test}}$ (i.e., a non-member). The attack dataset is thus defined as:
\begin{equation}
\mathcal{D}_{\text{Attack}} = \left\{ ( \mathbf{s}_v, y_v ) \middle \vert v \in \mathcal{G}_{\text{Shadow}}^{\text{Train}} \cup \mathcal{G}_{\text{Shadow}}^{\text{Test}} \right\}.
\end{equation}
Finally, we adopt a two-layer MLP as the attack model $\mathcal{F}_{\text{Attack}}$, trained on $\mathcal{D}_{\text{Attack}}$ using the cross-entropy loss. Once trained, the model can be used during inference to predict the membership status of a target node in the real target model $\mathcal{F}_{\text{target}}$ by its similarity-based attack feature vector.
The overall procedure and time complexity of our proposed \Model\ are summarized in Appendix~\ref{appendix:algorithm}.

\section{Experiments}

\subsection{Experimental Settings}

\subsubsection{Datasets.}
To evaluate the membership inference performance of \Model, we conduct experiments on five widely used benchmark datasets~\citet{zhao2024all}. Cora, CiteSeer, and PubMed~\cite{yang2016revisiting} are citation networks where nodes represent publications and edges denote citations. Computers and Photos~\cite{shchur2018pitfalls} are Amazon co-purchase graphs, where nodes are products and edges indicate frequent co-purchases.
Each dataset is randomly split into two equal halves: one for the target graph $\mathcal{G}_{\text{Target}}^{\text{Train}}$ and $\mathcal{G}_{\text{Target}}^{\text{Test}}$, respectively.
The target model is trained on $\mathcal{G}_{\text{Target}}^{\text{Train}}$, and membership inference treats nodes in $\mathcal{G}_{\text{Target}}^{\text{Train}}$ as members and nodes in $\mathcal{G}_{\text{Target}}^{\text{Test}}$ as non-members.

\subsubsection{Victims.} 
We evaluate the effectiveness of \Model\ against four multi-domain graph pre-trained models, categorized by their self-supervised learning objectives. 
(1) Contrastive learning: GCOPE~\cite{zhao2024all} introduces virtual nodes to connect different domains and dynamically optimizes them during training; SAMGPT~\cite{yu2025samgpt} enhances structural consistency across domains via structure tokens that unify message aggregation. 
(2) Link prediction: MDGPT~\cite{yu2024text} uses domain tokens to align semantic features across domains, while BRIDGE~\cite{yuan2025how} employs a domain aligner to capture shared patterns and suppress domain-specific noise.

\begin{table*}[!t]
  \centering
  \setlength{\tabcolsep}{0.5\tabcolsep}
  \resizebox{\textwidth}{!}{%
    \begin{tabular}{c|c|cc|cc|cc|cc|cc}
    \toprule
    \multicolumn{2}{c}{Dataset} & \multicolumn{2}{c|}{Cora} & \multicolumn{2}{c|}{CiteSeer} & \multicolumn{2}{c|}{PubMed} & \multicolumn{2}{c|}{Photo} & \multicolumn{2}{c}{Computers} \\
    \midrule
    Victims & Method & ACC   & F1    & ACC   & F1    & ACC   & F1    & ACC   & F1    & ACC   & F1 \\
    \midrule
    \multirow{7}[4]{*}{MDGPT} & Embed-MIA & 68.89\scalebox{0.65}{±1.83} & 60.31\scalebox{0.65}{±4.27} & 66.53\scalebox{0.65}{±1.26} & 53.25\scalebox{0.65}{±2.90} & 60.60\scalebox{0.65}{±0.80} & 61.93\scalebox{0.65}{±0.68} & 60.81\scalebox{0.65}{±2.52} & 64.99\scalebox{0.65}{±1.16} & 61.54\scalebox{0.65}{±0.51} & 64.70\scalebox{0.65}{±0.73} \\
          & Grad-MIA & 51.51\scalebox{0.65}{±1.41} & 22.03\scalebox{0.65}{±7.08} & 50.76\scalebox{0.65}{±0.49} & 14.29\scalebox{0.65}{±1.84} & 49.21\scalebox{0.65}{±2.79} & 35.29\scalebox{0.65}{±4.61} & 50.74\scalebox{0.65}{±5.16} & 16.65\scalebox{0.65}{±6.53} & 55.15\scalebox{0.65}{±4.12} & 41.17\scalebox{0.65}{±7.13} \\
          & NLO-MIA & 59.39\scalebox{0.65}{±0.69} & 50.04\scalebox{0.65}{±1.76} & 60.75\scalebox{0.65}{±0.74} & 53.42\scalebox{0.65}{±2.11} & 54.31\scalebox{0.65}{±0.52} & 54.51\scalebox{0.65}{±0.83} & 55.46\scalebox{0.65}{±2.76} & 54.70\scalebox{0.65}{±3.49} & 60.85\scalebox{0.65}{±3.15} & 61.55\scalebox{0.65}{±1.99} \\
          & GLO-MIA & 50.00\scalebox{0.65}{±0.00} & 66.67\scalebox{0.65}{±0.00} & 50.00\scalebox{0.65}{±0.00} & \underline{66.67}\scalebox{0.65}{±0.00} & 50.00\scalebox{0.65}{±0.00} & \underline{66.67}\scalebox{0.65}{±0.00} & 50.00\scalebox{0.65}{±0.00} & 66.67\scalebox{0.65}{±0.00} & 50.00\scalebox{0.65}{±0.00} & 66.67\scalebox{0.65}{±0.00} \\
          & GE-MIA & 60.79\scalebox{0.65}{±1.63} & 67.63\scalebox{0.65}{±1.97} & 54.90\scalebox{0.65}{±1.60} & 61.06\scalebox{0.65}{±1.99} & 51.69\scalebox{0.65}{±0.83} & 55.04\scalebox{0.65}{±3.70} & 53.34\scalebox{0.65}{±2.44} & 58.83\scalebox{0.65}{±7.07} & 53.16\scalebox{0.65}{±1.58} & 59.99\scalebox{0.65}{±3.41} \\
          & GPIA  & \underline{72.20}\scalebox{0.65}{±16.41} & \underline{76.41}\scalebox{0.65}{±15.07} & \underline{68.58}\scalebox{0.65}{±17.65} & 48.45\scalebox{0.65}{±38.29} & \underline{65.75}\scalebox{0.65}{±4.34} & 62.19\scalebox{0.65}{±15.12} & 6\underline{1.95}\scalebox{0.65}{±16.17} & \underline{73.13}\scalebox{0.65}{±8.84} & \underline{68.35}\scalebox{0.65}{±4.30} & \underline{65.84}\scalebox{0.65}{±15.79} \\
\cmidrule{2-12}          & \textbf{\Model} & \textbf{81.79}\scalebox{0.65}{±0.94} & \textbf{83.99}\scalebox{0.65}{±0.87} & \textbf{77.36}\scalebox{0.65}{±1.31} & \textbf{80.06}\scalebox{0.65}{±2.08} & \textbf{74.77}\scalebox{0.65}{±0.47} & \textbf{77.09}\scalebox{0.65}{±1.06} & \textbf{74.05}\scalebox{0.65}{±0.33} & \textbf{77.23}\scalebox{0.65}{±0.55} & \textbf{80.66}\scalebox{0.65}{±1.43} & \textbf{82.05}\scalebox{0.65}{±1.33} \\
    \midrule
    \multirow{7}[4]{*}{BRIDGE} & Embedding & 66.62\scalebox{0.65}{±1.02} & 58.93\scalebox{0.65}{±1.74} & \underline{65.61}\scalebox{0.65}{±1.50} & 52.31\scalebox{0.65}{±2.95} & 55.46\scalebox{0.65}{±0.48} & 58.65\scalebox{0.65}{±1.07} & \underline{59.43}\scalebox{0.65}{±1.15} & 64.47\scalebox{0.65}{±1.01} & \underline{58.94}\scalebox{0.65}{±1.59} & 64.20\scalebox{0.65}{±1.17} \\
          & Gradient & 49.91\scalebox{0.65}{±1.75} & 32.35\scalebox{0.65}{±2.33} & 50.00\scalebox{0.65}{±2.30} & 40.33\scalebox{0.65}{±11.27} & 51.45\scalebox{0.65}{±3.02} & 51.39\scalebox{0.65}{±3.68} & 49.06\scalebox{0.65}{±5.35} & 49.60\scalebox{0.65}{±3.89} & 45.44\scalebox{0.65}{±3.32} & 47.14\scalebox{0.65}{±3.13} \\
          & NLO-MIA & 60.97\scalebox{0.65}{±1.02} & 53.83\scalebox{0.65}{±1.53} & 61.48\scalebox{0.65}{±1.63} & 53.53\scalebox{0.65}{±2.28} & 52.17\scalebox{0.65}{±0.56} & 52.13\scalebox{0.65}{±1.05} & 53.38\scalebox{0.65}{±4.49} & 55.48\scalebox{0.65}{±4.72} & 54.54\scalebox{0.65}{±3.83} & 56.39\scalebox{0.65}{±3.23} \\
          & GLO-MIA & 50.92\scalebox{0.65}{±1.28} & 66.25\scalebox{0.65}{±0.93} & 50.52\scalebox{0.65}{±0.72} & \underline{66.06}\scalebox{0.65}{±0.94} & 49.98\scalebox{0.65}{±0.07} & \underline{66.62}\scalebox{0.65}{±0.12} & 50.02\scalebox{0.65}{±0.04} & \underline{65.47}\scalebox{0.65}{±2.67} & 48.16\scalebox{0.65}{±4.01} & \underline{66.63}\scalebox{0.65}{±0.08} \\
          & GE-MIA & 55.75\scalebox{0.65}{±3.43} & 59.34\scalebox{0.65}{±4.37} & 52.86\scalebox{0.65}{±2.36} & 52.77\scalebox{0.65}{±8.76} & 50.66\scalebox{0.65}{±0.30} & 52.13\scalebox{0.65}{±2.74} & 52.63\scalebox{0.65}{±1.22} & 62.41\scalebox{0.65}{±1.84} & 53.20\scalebox{0.65}{±1.85} & 57.58\scalebox{0.65}{±7.01} \\
          & GPIA  & \underline{66.76}\scalebox{0.65}{±11.87} & \underline{66.51}\scalebox{0.65}{±13.16} & 62.77\scalebox{0.65}{±15.71} & 46.22\scalebox{0.65}{±42.40} & \underline{59.34}\scalebox{0.65}{±5.59} & 55.36\scalebox{0.65}{±10.67} & 53.28\scalebox{0.65}{±5.64} & 47.08\scalebox{0.65}{±30.14} & 54.38\scalebox{0.65}{±4.59} & 38.15\scalebox{0.65}{±26.29} \\
\cmidrule{2-12}          & \textbf{\Model} & \textbf{81.20}\scalebox{0.65}{±1.10} & \textbf{79.97}\scalebox{0.65}{±1.26} & \textbf{79.57}\scalebox{0.65}{±1.25} & \textbf{80.94}\scalebox{0.65}{±1.46} & \textbf{74.93}\scalebox{0.65}{±0.69} & \textbf{79.05}\scalebox{0.65}{±0.28} & \textbf{70.36}\scalebox{0.65}{±1.74} & \textbf{73.06}\scalebox{0.65}{±2.75} & \textbf{73.39}\scalebox{0.65}{±0.43} & \textbf{76.13}\scalebox{0.65}{±1.05} \\
    \bottomrule
    \end{tabular}%
  }
 \caption{Membership inference attack performance against \textbf{link-prediction-based} multi-domain graph pre-trained models. Best results are in \textbf{bold}, and runner-ups are \underline{underlined}.}
  \label{tab:link_prediction}
\end{table*}

\begin{table*}[!t]
  \centering
  \setlength{\tabcolsep}{0.5\tabcolsep}
  \resizebox{\textwidth}{!}{%
    \begin{tabular}{c|c|cc|cc|cc|cc|cc}
    \toprule
    \multicolumn{2}{c}{Dataset} & \multicolumn{2}{c|}{Cora} & \multicolumn{2}{c|}{CiteSeer} & \multicolumn{2}{c|}{PubMed} & \multicolumn{2}{c|}{Photo} & \multicolumn{2}{c}{Computers} \\
    \midrule
    Victims & Method & ACC   & F1    & ACC   & F1    & ACC   & F1    & ACC   & F1    & ACC   & F1 \\
    \midrule
    \multirow{7}[4]{*}{GCOPE} & Embed-MIA & 60.00\scalebox{0.65}{±22.36} & \underline{73.33}\scalebox{0.65}{±14.91} & 50.00\scalebox{0.65}{±0.00} & \underline{66.67}\scalebox{0.65}{±0.00} & \underline{70.32}\scalebox{0.65}{±27.10} & \underline{76.73}\scalebox{0.65}{±22.27} & 60.00\scalebox{0.65}{±22.36} & 20.00\scalebox{0.65}{±44.72} & 70.00\scalebox{0.65}{±27.39} & 40.00\scalebox{0.65}{±54.77} \\
          & Grad-MIA & 48.42\scalebox{0.65}{±4.94} & 50.71\scalebox{0.65}{±7.92} & 51.02\scalebox{0.65}{±3.30} & 54.16\scalebox{0.65}{±0.47} & 57.81\scalebox{0.65}{±3.86} & 57.11\scalebox{0.65}{±6.99} & 52.47\scalebox{0.65}{±7.20} & 55.51\scalebox{0.65}{±7.53} & 57.97\scalebox{0.65}{±4.67} & 60.53\scalebox{0.65}{±2.22} \\
          & NLO-MIA & 54.58\scalebox{0.65}{±2.80} & 53.93\scalebox{0.65}{±3.64} & 53.97\scalebox{0.65}{±2.09} & 53.10\scalebox{0.65}{±1.54} & 51.94\scalebox{0.65}{±0.44} & 48.64\scalebox{0.65}{±1.83} & 55.21\scalebox{0.65}{±4.53} & 57.11\scalebox{0.65}{±4.39} & 55.22\scalebox{0.65}{±3.28} & 57.76\scalebox{0.65}{±6.51} \\
          & GLO-MIA & 41.89\scalebox{0.65}{±7.82} & 33.12\scalebox{0.65}{±27.55} & 44.98\scalebox{0.65}{±0.69} & 60.81\scalebox{0.65}{±1.17} & 48.64\scalebox{0.65}{±2.34} & 65.42\scalebox{0.65}{±2.16} & 45.46\scalebox{0.65}{±6.10} & \underline{62.30}\scalebox{0.65}{±6.06} & 49.87\scalebox{0.65}{±0.16} & 53.21\scalebox{0.65}{±29.75} \\
          & GE-MIA & 51.19\scalebox{0.65}{±1.36} & 55.32\scalebox{0.65}{±9.01} & 50.63\scalebox{0.65}{±0.55} & 38.08\scalebox{0.65}{±16.77} & 50.85\scalebox{0.65}{±0.44} & 32.04\scalebox{0.65}{±6.76} & 50.96\scalebox{0.65}{±0.94} & 35.96\scalebox{0.65}{±27.88} & 50.71\scalebox{0.65}{±0.30} & 16.62\scalebox{0.65}{±8.18} \\
          & GPIA  & \underline{80.00}\scalebox{0.65}{±27.39} & 60.00\scalebox{0.65}{±54.77} & \underline{70.00}\scalebox{0.65}{±27.39} & 66.67\scalebox{0.65}{±40.82} & 60.00\scalebox{0.65}{±41.83} & 40.00\scalebox{0.65}{±54.77} & \underline{80.00}\scalebox{0.65}{±27.39} & 60.00\scalebox{0.65}{±54.77} & \textbf{90.00}\scalebox{0.65}{±22.36} & \textbf{93.33}\scalebox{0.65}{±14.91} \\
\cmidrule{2-12}          & \textbf{\Model} & \textbf{87.21}\scalebox{0.65}{±1.04} & \textbf{88.13}\scalebox{0.65}{±0.83} & \textbf{85.86}\scalebox{0.65}{±0.75} & \textbf{87.44}\scalebox{0.65}{±0.60} & \textbf{80.20}\scalebox{0.65}{±1.77} & \textbf{83.37}\scalebox{0.65}{±1.17} & \textbf{83.19}\scalebox{0.65}{±1.05} & \textbf{85.04}\scalebox{0.65}{±0.69} & \underline{84.80}\scalebox{0.65}{±1.43} & \underline{86.35}\scalebox{0.65}{±0.84} \\
    \midrule
    \multirow{7}[4]{*}{SAMGPT} & Embedding & 54.39\scalebox{0.65}{±9.81} & 14.10\scalebox{0.65}{±31.53} & 51.18\scalebox{0.65}{±10.02} & 43.49\scalebox{0.65}{±33.11} & 48.61\scalebox{0.65}{±5.00} & 16.54\scalebox{0.65}{±29.76} & 50.17\scalebox{0.65}{±1.43} & 40.22\scalebox{0.65}{±36.62} & 49.80\scalebox{0.65}{±1.14} & 42.90\scalebox{0.65}{±32.76} \\
          & Gradient & 61.82\scalebox{0.65}{±2.33} & 60.71\scalebox{0.65}{±2.50} & 52.19\scalebox{0.65}{±2.19} & 47.71\scalebox{0.65}{±3.34} & 50.03\scalebox{0.65}{±1.22} & 51.76\scalebox{0.65}{±2.20} & 48.22\scalebox{0.65}{±1.82} & 52.86\scalebox{0.65}{±7.42} & 54.70\scalebox{0.65}{±3.70} & 61.21\scalebox{0.65}{±4.21} \\
          & NLO-MIA & 52.87\scalebox{0.65}{±1.97} & 52.66\scalebox{0.65}{±3.14} & 49.84\scalebox{0.65}{±2.58} & 49.63\scalebox{0.65}{±4.49} & 49.51\scalebox{0.65}{±0.18} & 51.07\scalebox{0.65}{±2.62} & 49.11\scalebox{0.65}{±4.85} & 49.68\scalebox{0.65}{±4.85} & 50.53\scalebox{0.65}{±2.80} & 50.67\scalebox{0.65}{±0.77} \\
          & GLO-MIA & 61.02\scalebox{0.65}{±0.61} & 63.52\scalebox{0.65}{±3.11} & 55.16\scalebox{0.65}{±5.31} & 47.16\scalebox{0.65}{±30.81} & 53.71\scalebox{0.65}{±2.09} & 59.74\scalebox{0.65}{±11.10} & 51.18\scalebox{0.65}{±1.93} & 63.45\scalebox{0.65}{±7.48} & 50.98\scalebox{0.65}{±1.23} & 35.32\scalebox{0.65}{±30.08} \\
          & GE-MIA & \underline{73.32}\scalebox{0.65}{±3.88} & \underline{74.99}\scalebox{0.65}{±4.52} & \underline{73.97}\scalebox{0.65}{±4.44} & \underline{75.67}\scalebox{0.65}{±5.64} & \underline{55.21}\scalebox{0.65}{±0.35} & 51.03\scalebox{0.65}{±6.16} & 50.61\scalebox{0.65}{±0.66} & 44.41\scalebox{0.65}{±11.64} & 55.77\scalebox{0.65}{±0.68} & 55.34\scalebox{0.65}{±3.39} \\
          & GPIA  & 58.55\scalebox{0.65}{±2.76} & 59.11\scalebox{0.65}{±7.01} & 55.31\scalebox{0.65}{±2.30} & 57.25\scalebox{0.65}{±5.80} & 54.59\scalebox{0.65}{±1.23} & \underline{61.83}\scalebox{0.65}{±3.24} & \underline{84.54}\scalebox{0.65}{±4.89} & \underline{83.94}\scalebox{0.65}{±3.92} & \underline{73.33}\scalebox{0.65}{±16.28} & \underline{77.20}\scalebox{0.65}{±11.04} \\
\cmidrule{2-12}          & \textbf{\Model} & \textbf{99.91}\scalebox{0.65}{±0.20} & \textbf{99.88}\scalebox{0.65}{±0.27} & \textbf{98.83}\scalebox{0.65}{±1.17} & \textbf{98.86}\scalebox{0.65}{±1.14} & \textbf{91.30}\scalebox{0.65}{±8.61} & \textbf{92.37}\scalebox{0.65}{±7.20} & \textbf{98.11}\scalebox{0.65}{±3.22} & \textbf{98.12}\scalebox{0.65}{±2.93} & \textbf{91.72}\scalebox{0.65}{±17.29} & \textbf{93.92}\scalebox{0.65}{±12.38} \\
    \bottomrule
    \end{tabular}%
  }
  \caption{Membership inference attack performance against \textbf{contrastive-learning-based} multi-domain graph pre-trained models. Best results are in \textbf{bold}, and runner-ups are \underline{underlined}.}
  \label{tab:contrastive_learning}
  \vspace{-0.5em}
\end{table*}

\subsubsection{Baselines.} 
We compare our proposed \Model\ with several related baseline methods. 
Since no existing approach specifically targets multi-domain graph pre-trained models, we adapt suitable variants of relevant baseline methods. Implementation details are provided in Appendix~\ref{appendix:expenriment_details}. The evaluated baselines include:
(1) Embed-MIA~\cite{duddu2020quantifying}, which trains an attack model directly on the output embeddings of the nodes in the shadow graph to distinguish members from non-members.
(2) Grad-MIA~\cite{nasr2019comprehensive}, which leverages the loss gradients with respect to the input node to train an attack model, based on the intuition that member samples produce smaller and more distinctive gradients.
(3) NLO-MIA~\cite{conti2022label}, which is a label-only attack that trains an attack model to infer membership based on embedding robustness under structural perturbations.
(4) GLO-MIA~\cite{dai2025graph}: which is a black-box attack that perturbs the input graph and infers membership by adopting a threshold over the output similarity.
(5) GE-MIA~\cite{duddu2020quantifying}, which assumes access to a few real member and non-member samples, and infers membership based on their distances to embedding cluster centers.
(6) GPIA~\cite{wang2022group}, which fine-tunes the model on individual samples to capture parameter changes for training an attack model, assuming access to a few real member and non-member samples.

\subsubsection{Metrics}
We adopt Accuracy (ACC) and F1-score (F1) as the evaluation metrics to assess the effectiveness of MIAs. 
Accuracy measures the overall proportion of correctly classified member and non-member samples, while F1-score provides a balanced evaluation of precision and recall.  

\subsubsection{Implements.} 
All the victims are pre-trained across five graph datasets. 
The attacker can only access a shadow graph with a similar distribution to the target node under inference and does not have access to graphs from all domains.
For the attacker model, both \Model\ and all baselines use a two-layer MLP with a latent dimension of 256. Baseline hyperparameters follow the settings recommended in their original papers and are further fine-tuned for fairness. 
Full hyperparameter configurations and implementation details are provided in Appendix~\ref{appendix:expenriment_details}. All experiments are conducted on a single NVIDIA V100 GPU and repeated 5 times.

\begin{figure*}[!t]
  \centering
  \begin{minipage}[t]{0.495\textwidth}
    \centering
    \includegraphics[width=\linewidth]{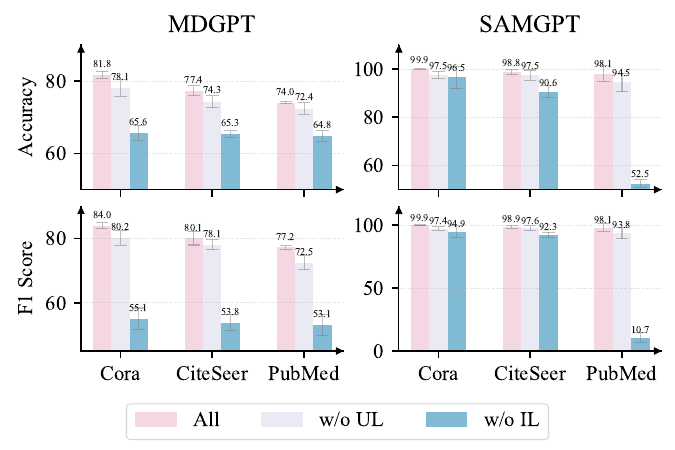}
    \caption{The ablation study of \Model.}
    \label{fig:ablation}
  \end{minipage}
  \begin{minipage}[t]{0.495\textwidth}
    \centering
    \includegraphics[width=\linewidth]{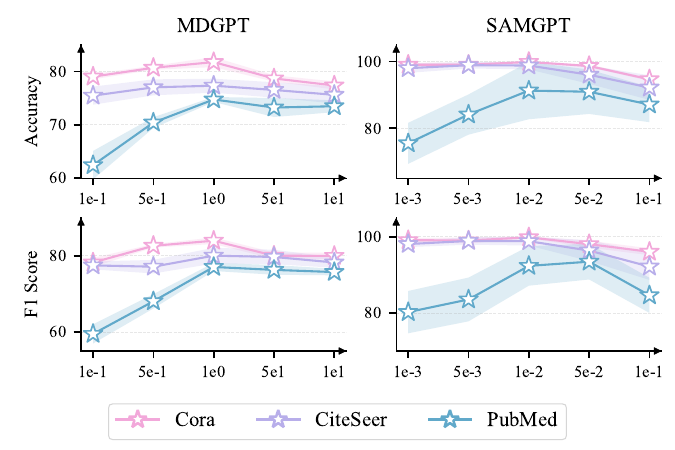}
    \caption{The study of hyperparameter $\alpha$}
    \label{fig:hyperparameter}
  \end{minipage}
\end{figure*}

\subsection{Attacking Link-Prediction-Based Multi-Domain Graph Pre-trained Models}
We evaluate the performance of \Model\ against two link-prediction–based multi-domain graph pre-trained models: MDGPT and BRIDGE. Membership inference results in terms of accuracy and F1 score are reported in Table~\ref{tab:link_prediction}.

\subsubsection{Analysis.} 
Table~\ref{tab:link_prediction} shows that our proposed \Model\ consistently achieves the best performance across all datasets and both link prediction-based multi-domain graph pretraining models, MDGPT and BRIDGE. Compared to the strongest baseline (GPIA), \Model\ improves accuracy and F1 by 9.6\% and 7.5\% respectively on Cora with MDGPT, even though GPIA benefits from access to real member and non-member samples. Similar improvements are observed on other datasets, with accuracy and F1 gains reaching up to 10.8\% and 16.3\%. 
These results highlight the effectiveness of \Model\ even using only output embeddings, without requiring privileged supervision.

\subsection{Attacking Contrastive-Learning-Based Multi-Domain Graph Pretrained Models}
We evaluate the performance of our proposed \Model\ against the contrastive-learning-based multi-domain graph pre-trained models, specifically GCOPE and SAMGPT. Membership inference results in terms of accuracy and F1 score are reported in Table~\ref{tab:contrastive_learning}.

\subsubsection{Analysis.}
As shown in Table~\ref{tab:contrastive_learning}, \Model\ consistently outperforms all baselines across most datasets for both GCOPE and SAMGPT. 
On GCOPE, it achieves the highest accuracy and F1 in all cases except the Computers dataset, where GPIA slightly leads. However, GPIA assumes access to some real member and non-member nodes, while \Model\ requires no such privileged information. 
On SAMGPT, \Model\ achieves the best performance across all datasets. For example, it improves over the closest baseline by 26.6\% in accuracy and 25.0\% in F1 on Cora, and by over 20\% in both metrics on Computers.
These results highlight the effectiveness and generality of \Model\ in attacking contrastive-learning-based pre-trained models under more practical and constrained settings.

\subsection{Ablation Study}
\label{sec:experiment_ablation}
To evaluate the contribution of each component in our proposed \Model, we construct two ablated variants:

\begin{itemize}
    \item \textbf{\Model~(w/o UL):} This variant removes the Membership Signal Amplification mechanism described in Section~\ref{sec:unlearning}, which is designed to enhance membership signals. Instead, it fine-tunes the target model directly to obtain the shadow model, omitting the unlearning step.
    \item \textbf{\Model~(w/o IL):} This variant removes the Incremental Shadow Model Construction mechanism, which aims to replicate the overfitting behavior of the target model using limited shadow data. Instead, it trains the shadow model from scratch on the shadow dataset.
\end{itemize}
We select MDGPT, which is based on the link prediction objective, and SAMGPT, which is based on contrastive learning, as the target victims. 
Experiments are conducted on the Cora, CiteSeer, and PubMed datasets. The results of the ablation study are shown in Figure~\ref{fig:ablation}. As illustrated, both components contribute to the overall performance of \Model. The incremental learning module enables the shadow model to more closely replicate the overfitting behavior of the target model, providing substantial intrinsic gains. The unlearning component further improves performance by amplifying membership signals, offering additional benefits.

\subsection{Hyperparameter Study}
In this section, we analyze the sensitivity of \Model\ to the hyperparameter $\alpha$, which controls the regularization strength during the incremental learning stage by constraining important parameters identified by the Fisher Information Matrix in Eq.~\eqref{eq:fisher}. Using the same victim models and dataset settings as the ablation analysis in Section~\ref{sec:experiment_ablation}, the results are presented in Figure~\ref{fig:hyperparameter}. We observe that \Model\ is relatively robust to changes in $\alpha$, maintaining stable performance across a wide range of values. Moreover, the optimal setting remains nearly the same across different datasets.

\section{Conclusion}

This paper investigates the privacy risks of multi-domain graph pre-trained models via Membership Inference Attacks (MIAs). We show that existing graph-based MIA methods are largely ineffective in this setting, as output embeddings contain limited overfitting signals.
To address this, we propose \Model, a novel MIA framework tailored for multi-domain graph pre-training. \Model\ amplifies membership signals through unlearning to counteract the generalization from multi-domain training, adopts an incremental shadow model construction strategy to replicate the target model’s overfitting behavior with limited shadow data, and infers membership by measuring similarity to positive and negative samples.
Extensive experiments on representative models demonstrate the effectiveness of \Model\ and reveal the privacy risks of multi-domain graph pre-training.

\clearpage
\section*{Acknowledgements}
The corresponding author is Jianxin Li.
This work was supported by the National Natural Science Foundation of China under Grants No. 62225202 and No. 62302023, and by the Fundamental Research Funds for the Central Universities.
We express our sincere gratitude to all reviewers for their valuable efforts and contributions.
\bibliography{reference}

\clearpage
\setcounter{secnumdepth}{2}
\appendix
\section{Time Complexity Analysis}
\label{appendix:compelxity}
In this section, we present the complexity analysis of the proposed \Model. The overall time complexity is primarily determined by the following three components:
\begin{itemize}
    \item \textit{Membership Signal Amplification Mechanism}: This process involves fine-tuning the model on the unlearning graph $\mathcal{G}_{\text{Unlearn}}$, incurring a time complexity of $\mathcal{O}\left(E_{1} n_{\text{unlearn}} Lfd\right)$, where $n_{\text{unlearn}}$ denotes the number of nodes in $\mathcal{G}_{\text{Unlearn}}$, $E_{1}$ is the number of fine-tuning epochs, $L$ is the number of GNN layers, $f$ is the average neighborhood size, and $d$ is the embedding dimension.
    \item \textit{Incremental Shadow Model Construction Mechanism}: This step fine-tunes shadow models on the shadow training graph $\mathcal{G}_{\text{Shadow}}^{\text{Train}}$, with a comparable cost of $\mathcal{O}\left(E_{2} n_{\text{train}} Lfd\right)$, where $n_{\text{train}}$ is the number of nodes in $\mathcal{G}_{\text{Shadow}}^{\text{Train}}$ and $E_{2}$ is the number of fine-tuning epochs.
    \item \textit{Similarity-Based Inference Mechanism}: This stage requires only a single forward pass for each node and its sampled neighbors, resulting in a complexity of $\mathcal{O}(mLfd)$, where $m$ is the number of queried nodes used for similarity inference.
\end{itemize}
Hence, the total time complexity of our proposed \Model\ is $\mathcal{O}\left((E_1 n_{\text{unlearn}} + E_2 n_{\text{train}} + m)Lfd\right)$, which scales linearly with the graph size and embedding dimension.

\section{Algorithm}\label{appendix:algorithm}
The overall procedure of our proposed \Model\ is summarized in Algorithm~\ref{algorithm:model}. 
\begin{algorithm}
\caption{Overall pipeline of \Model.}
\label{algorithm:model}
\KwIn {Shadow graph $\mathcal{G}_{\text{Shadow}}$; Target Model $\mathcal{F}_{\text{Target}}$; Hyperparameter $\alpha$.}
\KwOut{Attack model $\mathcal{F}_{\text{Attack}}$.}
\BlankLine 

Randomly partition the shadow graph $\mathcal{G}_{\text{Shadow}}$ into $\mathcal{G}_{\text{Unlearn}}$, $\mathcal{G}_{\text{Shadow}}^{\text{Train}}$ and $\mathcal{G}_{\text{Shadow}}^{\text{Test}}$ \;

Amplify membership signals by applying machine unlearning on $\mathcal{G}_{\text{Unlearn}}$ to obtain the unlearned model $\mathcal{F}_{\text{Unlearn}}$ $\leftarrow$ (Section~\ref{sec:unlearning})\;

Construct the shadow model $\mathcal{F}_{\text{Shadow}}$ incrementally based on $\mathcal{G}_{\text{Shadow}}^{\text{Train}}$ $\leftarrow$ (Section~\ref{sec:incremental})\;

Build the similarity-based attack dataset $\mathcal{D}_{\text{Attack}}$ and obtain the attack model $\mathcal{F}_{\text{Attack}}$  $\leftarrow$ (Section~\ref{sec:similarity}).

\end{algorithm}

\section{Pre-attack Analysis Details}
\label{appendix:pre_attack}
The selected multi-domain graph pre-trained models (SAMGPT and MDGPT) are trained on five datasets, each associated with a specific target training graph $\mathcal{G}_{\text{Target}}^{\text{Train}}$: Cora, CiteSeer, PubMed~\cite{yang2016revisiting}, Photo, and Computers~\cite{shchur2018pitfalls}. For evaluation, we directly analyze the output node embeddings of both the training graph $\mathcal{G}_{\text{Target}}^{\text{Train}}$ and the test graph $\mathcal{G}_{\text{Target}}^{\text{Test}}$ to examine whether existing graph MIAs can perform effective attacks against the multi-domain graph pre-trained models.

\begin{figure}[!t]
  \centering

  \begin{subfigure}[t]{0.8\linewidth}
    \centering
    \includegraphics[width=\linewidth]{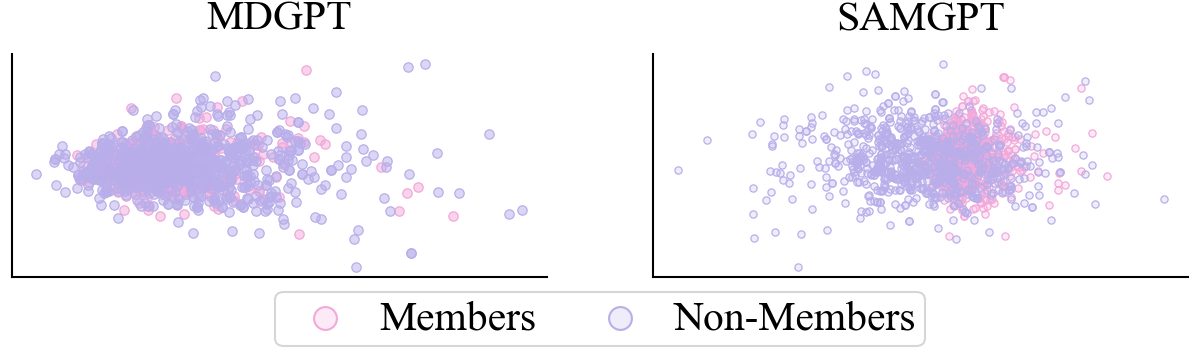}
    \caption{Separability analysis in CiteSeer.}
  \end{subfigure}

  \begin{subfigure}[t]{0.8\linewidth}
    \centering
    \includegraphics[width=\linewidth]{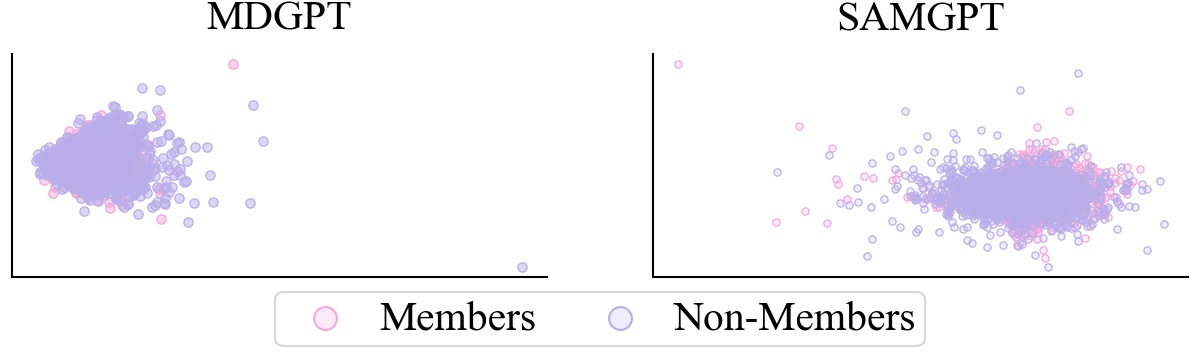}
    \caption{Separability analysis in PubMed.}
  \end{subfigure}

  \begin{subfigure}[t]{0.8\linewidth}
    \centering
    \includegraphics[width=\linewidth]{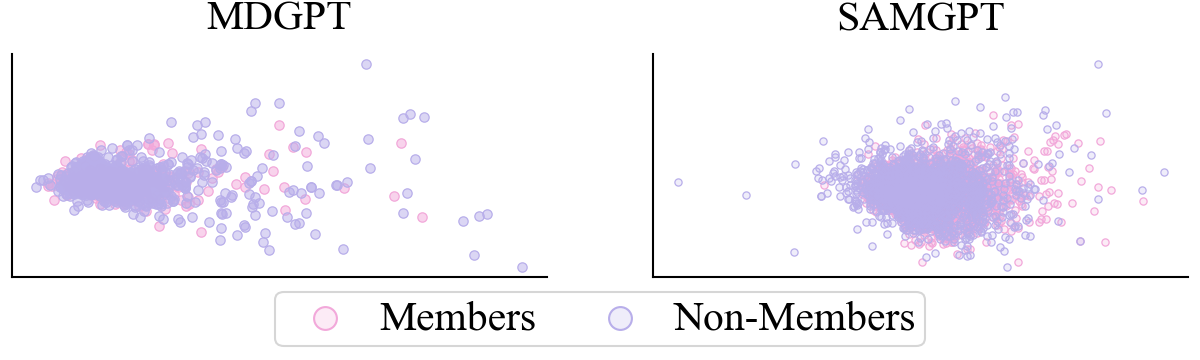}
    \caption{Separability analysis in Photo.}
  \end{subfigure}

  \begin{subfigure}[t]{0.8\linewidth}
    \centering
    \includegraphics[width=\linewidth]{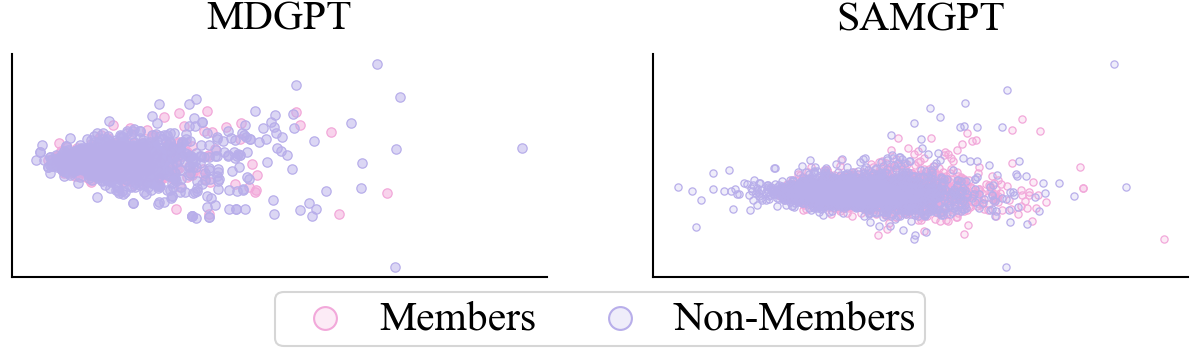}
    \caption{Separability analysis in Computers.}
  \end{subfigure}

  \caption{Additional separability analysis of embeddings.}
  \label{fig:appendix_separability}
\end{figure}

In the first toy experiment, we examine whether the output node embeddings possess class separability properties similar to those of logits. To this end, we apply Principal Component Analysis (PCA)~\cite{abdi2010principal} to reduce the embedding dimensionality and visualize the results to assess whether member and non-member embeddings are distinguishable. 
Additional results for CiteSeer, PubMed, Photo, and Computers are presented in Figure~\ref{fig:appendix_separability}.

\begin{figure}[!t]
  \centering

  \begin{subfigure}[t]{0.8\linewidth}
    \centering
    \includegraphics[width=\linewidth]{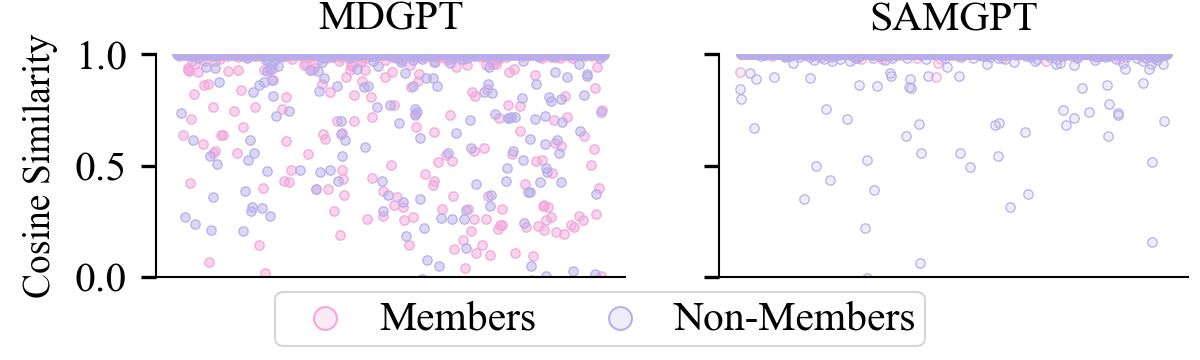}
    \caption{Robustness analysis in CiteSeer.}
  \end{subfigure}

  \begin{subfigure}[t]{0.8\linewidth}
    \centering
    \includegraphics[width=\linewidth]{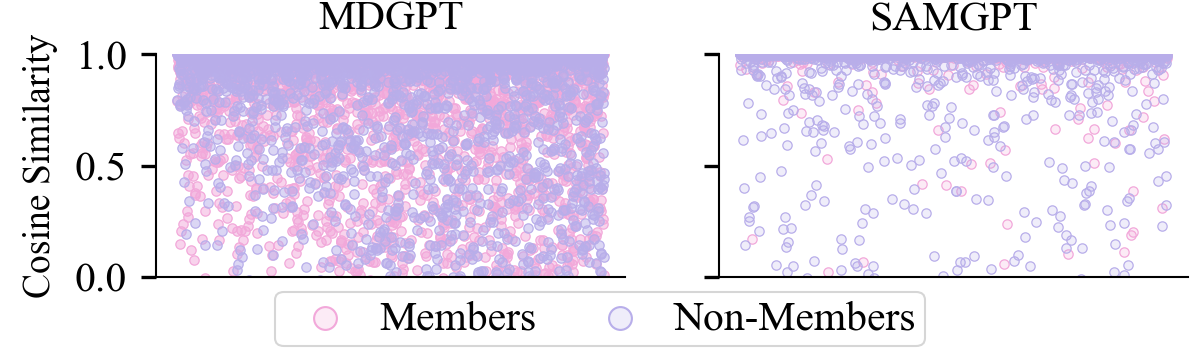}
    \caption{Robustness analysis in PubMed.}
  \end{subfigure}

  \begin{subfigure}[t]{0.8\linewidth}
    \centering
    \includegraphics[width=\linewidth]{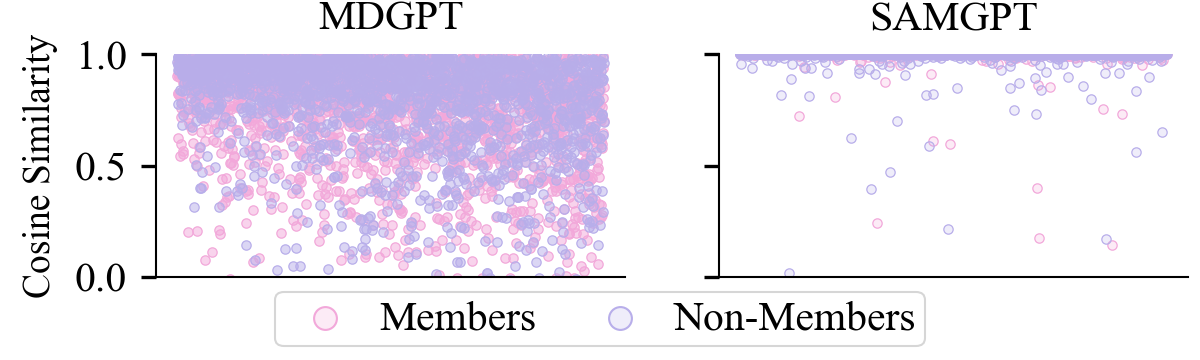}
    \caption{Robustness analysis in Photo.}
  \end{subfigure}

  \begin{subfigure}[t]{0.8\linewidth}
    \centering
    \includegraphics[width=\linewidth]{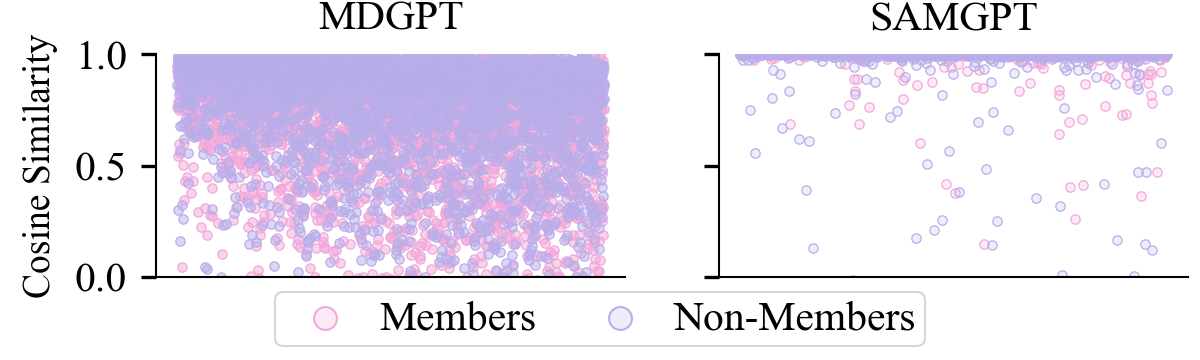}
    \caption{Robustness analysis in Computers.}
  \end{subfigure}

  \caption{Additional robustness analysis of embeddings.}
  \vspace{-0.5em}
  \label{fig:appendix_robustness}
\end{figure}

In the second toy experiment, we investigate whether the output node embeddings exhibit greater robustness for member nodes under input graph perturbations. 
Specifically, we apply random structural perturbations to the graph by adding or deleting edges, with a perturbation budget set to 15\% of the total number of graph edges. We then measure the output similarity of the node embeddings before and after perturbation.
Additional results for CiteSeer, PubMed, Photo, and Computers are shown in Figure~\ref{fig:appendix_robustness}.

Overall, we observe that the embeddings produced by the multi-domain graph pre-trained models do not demonstrate strong separability between member and non-member nodes, nor do they show increased robustness for members. This finding suggests that existing graph-based MIA methods, which rely on membership signals such as the confidence of model logits or the robustness of model outputs, are ineffective in scenarios involving multi-domain graph pre-trained models.

\section{Experiment Details}
\label{appendix:expenriment_details}
In this section, we elaborate on the experimental details presented in this paper, including information on the datasets, baselines, and implementation settings.

\subsection{Dataset Details}
Following prior work~\cite{zhao2024all}, we pre-train the models across five datasets, treating each graph as a separate domain. The five datasets are listed below, and their detailed statistics are presented in Table~\ref{tab:dataset_statistics}. All the datasets can be obtained in the PyG Python package.

\begin{itemize}
    \item \textit{Cora}~\cite{yang2016revisiting}: A citation network with 2,708 papers categorized into seven classes and connected by 5,429 directed citation links. Each paper is represented by a binary word vector over a dictionary of 1,433 unique words.
    \item \textit{CiteSeer}~\cite{yang2016revisiting}: Contains 3,312 publications classified into six categories, linked by 4,732 citation edges. Each document is described by a binary word vector over 3,703 unique terms.
    \item \textit{PubMed}~\cite{yang2016revisiting}: Comprises 19,717 diabetes-related publications from the PubMed database, categorized into three classes.
    \item \textit{Photo}~\cite{shchur2018pitfalls}: An Amazon product co-purchase graph with 7,650 nodes and 238,163 edges. Each node is encoded by a 745-dimensional bag-of-words feature vector from product reviews and labeled into one of eight categories.
    \item \textit{Computers}~\cite{shchur2018pitfalls}: An Amazon co-purchase graph with 13,752 nodes and 245,861 edges (491,722 after bidirectional deduplication). Each node is represented by a 767-dimensional bag-of-words vector and assigned to one of ten classes.
\end{itemize}
\begin{table}[!htbp]
  \centering
  \caption{Statistics of datasets.}
  \setlength{\tabcolsep}{0.5\tabcolsep}
  \resizebox{\linewidth}{!}{%
    \begin{tabular}{c|ccccc}
    \toprule
    Dataset & Cora  & CiteSeer & PubMed & Photo & Computers \\
    \midrule
    \# Nodes & 2,708 & 3,327 & 19,717 & 7,650 & 13,752 \\
    \# Edges & 10,556 & 9,104 & 88,648 & 238,162 & 491,722 \\
    \# Features & 1,433 & 3,703 & 500   & 745   & 767 \\
    \# Avg. Degree & 3.89  & 2.73  & 4.49  & 31.13 & 35.75 \\
    \bottomrule
    \end{tabular}%
  }
  \label{tab:dataset_statistics}
\end{table}

\subsection{Baseline Details}
Since no existing graph MIA method is specifically designed for multi-domain graph pre-trained models, we compare our proposed \Model\ against adapted variants of existing graph MIA approaches, as described below:
\begin{itemize}
    \item \textit{Embedding}~\cite{duddu2020quantifying}, trains a two-layer MLP attack model directly on the output embeddings of nodes produced by a shadow model trained on the shadow training graph $\mathcal{G}_{\text{Shadow}}^{\text{Train}}$, aiming to distinguish members from non-members.
    \item \textit{Gradient}~\cite{nasr2019comprehensive}, which leverages the loss gradients of a shadow model trained on $\mathcal{G}_{\text{Shadow}}^{\text{Train}}$, with respect to inputs from both $\mathcal{G}_{\text{Shadow}}^{\text{Train}}$ and $\mathcal{G}_{\text{Shadow}}^{\text{Test}}$, to train a two-layer MLP attack model. This method is based on the intuition that member samples produce smaller and more distinctive gradients.
    \item \textit{NLO-MIA}~\cite{conti2022label}, a label-only attack that trains a two-layer MLP to infer membership based on the robustness of embeddings under structural perturbations. Specifically, it randomly perturbs the graph structure by modifying 0.15\% of the edges in the input graph, repeated 10 times, and records the pairwise similarity between perturbed embeddings to form a 45-dimensional similarity vector used to train the attack model.
    \item \textit{GLO-MIA}~\cite{dai2025graph}: an attack that perturbs the input graph in the same way as NLO-MIA. However, instead of training a two-layer MLP attack model, it uses the average embedding similarity as the distinguishing characteristic and infers membership by applying a threshold over the output embedding similarity.
    \item \textit{GE-MIA}~\cite{duddu2020quantifying}, which assumes access to 20 real member and 20 non-member samples, and infers the membership of a given node based on its distance to the cluster centers of these reference embeddings, using a threshold-based decision rule.
    \item \textit{GPIA}~\cite{wang2022group} first trains a shadow model on $\mathcal{G}_{\text{Shadow}}^{\text{Train}}$, and then fine-tunes it on individual samples for 10 epochs to capture parameter change vectors. These vectors, collected from both $\mathcal{G}_{\text{Shadow}}^{\text{Train}}$ and $\mathcal{G}_{\text{Shadow}}^{\text{Test}}$, are used to train an attack model. During inference, the target model is fine-tuned on the queried sample, and the resulting parameter change vector is passed to the attack model to determine membership. This method assumes access to a small number of real member and non-member samples, and relies on the intuition that member samples cause smaller parameter changes than non-members.
\end{itemize}

\subsection{Implement Details}

\subsubsection{Hyperparameter Settings.} 
For the hyperparameter $\lambda$, we set $\lambda = 1$ in all experiments by default. For the hyperparameter $\alpha$, we set $\alpha = 1$ for MDGPT and BRIDGE across all datasets, and $\alpha = 1e^{-2}$ for GCOPE and SAMGPT, following preliminary tuning. Additional implementation details can be found in our code, which is included in the supplementary material.
\subsubsection{Running Environment.}
All experiments were conducted on a machine running Ubuntu 20.04 LTS. The system was equipped with an Intel(R) Xeon(R) Platinum 8358 CPU @ 2.60GHz and 1TB DDR4 memory. We used an NVIDIA Tesla V100 GPU with 32GB of memory for acceleration. The software environment included CUDA 12.4, Python 3.11.0, PyTorch 2.4.0, and PyTorch Geometric 2.6.1.

\end{document}